# Trustworthy AI

Jeannette M. Wing
Avanessians Director of the Data Science Institute and Professor of Computer Science
Columbia University

14 February 2020

For certain tasks, AI systems have achieved good enough performance to be deployed in our streets and our homes. Object recognition helps modern cars see. Speech recognition helps personalized voice assistants, such as Siri and Alexa, converse. For other tasks, AI systems have even exceeded human performance. AlphaGo was the first computer program to beat the best Go player in the world.

The promise of AI is huge. They will drive our cars. They will help doctors diagnose disease more accurately [Tiwari et al. 2016]. They will help judges make more consistent court decisions. They will help employers hire more suitable job candidates.

We know, however, these AI systems can be brittle and unfair. Adding graffiti to a stop sign fools the classifier into saying it is not a stop sign [Eykholt et al. 2017]. Adding noise to an image of a benign skin lesion fools the classifier into saying it is malignant [Finlayson et al 2019]. Risk assessment tools used in US courts have shown to be biased against blacks [Angwin et al. 2016]. Corporate recruiting tools have been shown to be biased against women [Dastin 2018].

How then can we deliver on the promise of the benefits of AI but address these scenarios that have life-critical consequences for people and society? In short, how can we achieve *trustworthy AI*?

## 1. From Trustworthy Computing to Trustworthy AI

The landmark *Trust in Cyberspace* 1999 National Academies report lay the foundations of trustworthy computing and what continues to be an active research area [NRC 1999].

The National Science Foundation started a series of programs on trust. Starting with Trusted Computing (initiated in 2001), then Cyber Trust (2004), then Trustworthy Computing (2007), and now Secure and Trustworthy Systems (2011), the Computer and Information Science and Engineering Directorate has grown the academic research community in trustworthy computing. Although it started within the computer science community, support for research in trustworthy computing now spans multiple directorates at NSF and engages many other funding organizations, including, through the Networking and Information Technology Research and Development (NITRD) Program, 20 federal agencies.

Industry has also been a leader and active participant in trustworthy computing. With Bill Gates's January 2002 "Trustworthy Computing" memo [Gates 2002], Microsoft signaled to its employees, customers, shareholders, and the rest of the information technology sector the importance of trustworthy software and hardware products. It referred to an internal Microsoft white paper, which identified four pillars to trustworthiness: security, privacy, reliability, and business integrity. The first three properties were aimed at the customer, to give the customer good reason to trust Microsoft software and services.

After two decades of investment and advances in research and development, *trustworthy* has come to mean a set of (overlapping) properties:

- Reliability: Does the system do the right thing?
- Safety: Does the system do no harm?
- Security: How vulnerable is the system to attack?
- Privacy: Does the system protect a person's identity and data?
- Availability: Is the system up when I need to access it?
- Usability: Can a human use it easily?

The *computing* systems for which we want such properties to hold are hardware and software systems, including their interaction with the humans and the physical world. As technology advances and as adversaries get more sophisticated, trustworthy computing remains a holy grail.

AI systems raise the bar in terms of the set of properties of interest. In addition to the properties associated with trustworthy computing (from above), we also want (overlapping) properties such as:

- Accuracy: How well does the AI system do on new (unseen) data compared to data on which it was trained and tested?
- Robustness: How sensitive is the system's outcome to a change in the input?
- Fairness: Are the system outcomes unbiased?
- Accountability: Who or what is responsible for the system's outcome?
- Transparency: Is it clear to an external observer how the system's outcome was produced?
- Interpretability/Explainability: Can the system's outcome be justified with an explanation that a human can understand and/or that is meaningful to the end user?
- Ethical: Was the data collected in an ethical manner? Will the system's outcome be used in an ethical manner?
- …and others, yet to be identified

The machine learning community considers accuracy as a gold standard, but trustworthy AI requires us to explore tradeoffs among these properties. For example, perhaps we are willing to give up on some accuracy in order to deploy a fairer model. Also, some of the above properties may have different interpretations, leading to different formalizations. For example, there are many reasonable notions of fairness [Narayanan 2018], e.g., group fairness (statistical parity) and individual fairness [Dwork et al. 2012], some of which are incompatible with each other [Chouldechova 2016, Kleinberg, Mullainathan, and Raghavan 2017].

Traditional software and hardware systems are complex due to their size and the number of interactions among their components. For the most part, we can define their behavior in terms of discrete logic and as deterministic state machines.

Today's AI systems, especially those using deep neural networks, add a dimension of complexity to traditional computing systems. This complexity is due to their inherent probabilistic nature. Through probabilities, AI systems model the uncertainty of human behavior and the uncertainty of the physical world. More recent advances in machine learning, which rely on big data, add to their probabilistic nature, as data from the real world are just points in a probability space. Thus, trustworthy AI necessarily directs our attention from the primarily deterministic nature of traditional computing systems to the probabilistic nature of AI systems.

## 2. Verify, to Trust

How can we design, implement, and deploy AI systems to be trustworthy?

One approach for building end-user trust in computing systems is formal verification, where properties are proven once and for all over a large domain, e.g., for all inputs to a program or for all behaviors of a concurrent or distributed system. Alternatively, the verification process identifies a counterexample, e.g., an input value where the program produces the wrong output or a behavior that fails to satisfy the desired property, and thus provides valuable feedback on how to improve the system. Formal verification has the advantage of obviating the need to test individual input values or behaviors one-by-one, which for large (or infinite) state spaces is impossible to achieve completely. These approaches are now used in the hardware and software industry, e.g., Intel [Harrison 2003], IBM [Baumbartner 2006], Microsoft [Ball et al. 2004], and Amazon [Newcombe et al. 2015]. Due to advances in formal methods languages, algorithms, and tools, and to the increased scale and complexity of hardware and software, we have seen in the past few years a new surge of interest and excitement in formal verification, especially for ensuring the correctness of critical components of system infrastructure [Bhargavan et al. 2017, Chen et al. 2015, Chen et al. 2017, Gu et al. 2016, Hawblitzel et al. 2014, Koh et al. 2019, Protzenko et al. 2017].

Formal verification is a way to provide provable guarantees and thus increase one's trust that the system will behave as desired.

### 2.1. From Traditional Formal Methods to Formal Methods for AI

In traditional formal methods, we want to show that a model M *satisfies* ($\models$) a property P.

$$M \models P$$

M is the object to be verified—be it a program or an abstract model of a complex system, e.g., a concurrent, distributed, or reactive system. P is the correctness property, expressed in some discrete logic. For example, M might be a concurrent program that uses locks for synchronization and P might be "deadlock free." A proof that M is deadlock free means any user of M is assured that M will never reach a deadlocked state. To prove that M satisfies P, we use formal mathematical logics, which are the basis of today's scalable and practical verification tools such as model checkers and satisfiability modulo theories (SMT) solvers.

Especially when M is a concurrent, distributed, or reactive system, in traditional formal methods, we often add explicitly a specification of a system's environment E in the formulation of the verification task:

$$E, M \vDash P$$

For example, if M is a parallel process, E might be another process with which M interacts (and then we might write $E \parallel M \vDash P$). Or, if M is device driver code, E might be a model of the operating system. Or, if M is a control system, E might be a model of its environment that closes the control loop. The specification of E is written to make explicit the assumptions about the environment in which the system is to be verified.

For verifying AI systems, M could be interpreted to be a complex system, e.g., a self-driving car, which has a component within it that is a machine-learned model, e.g., a computer vision system. Here, we would want to prove P, e.g., safety or robustness, with respect to M (the car) in the context of E (traffic, roads, pedestrians, buildings, and so on). We can view proving P as proving a "system-level" property. Seshia et al. elaborate on the formal specification challenges with this perspective [Seshia et al. 2018], where a deep neural network might be a component of the system M.

But what can we assert about the machine learned model, e.g., the DNN, that is a critical component of this system? Is there a robustness or fairness property we can verify of the machine-learned model itself? Answering these questions raises new verification challenges.

## 2.2. Verifying a Machine-Learned Model M

For verifying an ML model, we reinterpret M and P: M stands for a machine-learned model. P stands for a trustworthy property, e.g., safety, robustness, privacy, or fairness.

Verifying AI systems ups the ante over traditional formal methods. There are two key differences:

1. *The inherent probabilistic nature of M and P, and thus the need for probabilistic reasoning* ($\vDash$).

    - The ML model, M, itself is semantically and structurally different from a typical computer program. As mentioned, it is inherently probabilistic, taking inputs from the real world, that are perhaps mathematically modeled as a stochastic process, and producing outputs that are associated with probabilities. Internally, the model itself operates over probabilities; for example, labels on edges in a deep neural network are probabilities and nodes compute functions over these probabilities. Structurally, because a machine generated the ML model, M itself is not necessarily something human readable or comprehensible; crudely, a DNN is a complex structure of if-then-else statements that would unlikely ever be written by a human. This "intermediate code" representation opens up new lines of research in program analysis.

- The properties P themselves may be formulated over continuous, not (just) discrete domains, and/or using expressions from probability and statistics. Robustness properties for deep neural networks are characterized as predicates over continuous variables [Dreossi et al. 2019]. Fairness properties are characterized in terms of expectations with respect to a loss function over reals (e.g., see [Dwork et al. 2012]). Differential privacy is defined in terms of a difference in probabilities with respect to a (small) real value [Dwork et al. 2006]. Note that just as with properties such as usability for trustworthy computing, some desired properties of trustworthy AI systems, e.g., transparency or ethics, have yet to be formalized or may not be formalizable. Thus, verification of AI systems will be limited to what can be formalized.

- These inherently probabilistic models M and associated desired trust properties P call for scalable and/or new verification techniques that work over reals, non-linear functions, probability distributions, stochastic processes, and so on. Thus, one stepping stone to verifying AI systems is probabilistic logics and hybrid logics, used by the cyber-physical systems community. Even more challenging is that these verification techniques need to operate over machine-generated code, in particular code that itself might not be produced deterministically.[1]

*2. The role of data*

Perhaps the more significant key difference between traditional formal verification and verification for AI systems is the role of data—data used in training, testing, and deploying ML models. Today's ML models are built and used with respect to a set, D, of data. For verifying an ML model, we propose to make explicit the assumptions about this data, and formulate the verification problem as:

$$D, M \vDash P$$

Data is divided into *available data* and *unseen data*, where *available data* is data-at-hand, used for training and testing M; and *unseen data* is data over which M needs (or is expected) to operate without having seen it before. The whole idea behind building M is so that based on the data on which it was trained and tested, M would be able to make predictions on data it has never seen before, typically to some degree of accuracy.

Making the role of data explicit raises novel specification and verification questions, roughly broken into these categories:

<u>Collection and partitioning of available data</u>

- How do we partition an available (given) dataset into a training set and a test set? What guarantees can we make of this partition with respect to a desired property P, in building a model M?

---

[1] The ways in which machine learning models, some with millions of parameters, are constructed today, perhaps through weeks of training on clusters of CPUs, TPUs, and GPUs, raise a meta-issue of trust: scientific reproducibility.

- How much data suffices to build a model M for a given property P? Does adding more data to train or test M make it more robust, fairer, etc. or does it not have an effect with respect to the property P? What new kind of data needs to be collected if a desired property does not hold?

Specifying unseen data

- How do we specify the data and/or characterize properties of the data? For example, we could specify D as a stochastic process that generates inputs over which the ML model needs to be verified. Or, we could specify D as a data distribution. For a common statistical model, e.g., a normal distribution, we could specify D in terms of its parameters, e.g., mean and variance. Probabilistic programming languages, e.g., Stan [Carpenter et al. 2017], might be a starting point for specifying statistical models. But what of large real-world datasets that do not fit common statistical models or which have thousands of parameters?
- In specifying unseen data, by definition, we will need to make certain assumptions about the unseen data. Would these assumptions not then be the same as those we would make to build the model M in the first place? More to the point: *How can we trust the specification of D?* This seemingly logical deadlock is analogous to the problem in traditional verification, where given an M, we need to assume that the specifications of the elements E and P are "correct" in the verification task E, M ⊨ P. Then in the verification process, we may need to modify E and/or P (or even M). To break the circular reasoning at hand, one approach is to use a different validation approach for checking the specification of D; such approaches could borrow from a repertoire of statistical tools (see Section 2.3). Another approach would be to assume that an initial specification is small or simple enough that it can be checked by (say, manual) inspection; then we use this specification to bootstrap an iterative refinement process. (We draw inspiration from the counter-example guided abstraction and refinement method [Clarke et al. 2000] of formal methods.) This refinement process may necessitate modifying D, M, and/or P.
- How does the specification of unseen data relate to the specification of the data on which M was trained and tested?

In traditional verification, we aim to prove property, P, a universally quantified statement: for example, *for all* input values of integer variable *x*, the program will return a positive integer; or *for all* execution sequences *x*, the system will not deadlock.

So the first question for proving P of an ML model, M, is: in P, what do we quantify over? For an ML model that is to be deployed in the real world, one reasonable answer is to quantify over data distributions. But a ML model is meant to work only for certain distributions that are formed by real world phenomena, and *not* for arbitrary distributions. We do not want to prove a property *for all* data distributions. This insight on the difference in what we quantify over and what the data represents for proving a trust property for M leads to novel specification questions:

- How can we specify the class of distributions over which P should hold for a given M? Consider robustness and fairness as two examples:

- For robustness, in the adversarial machine learning setting, we might want to show that M is robust to all norm-bounded perturbations D. More interestingly, we might want to show M is robust to all "semantic" or "structural" perturbations for the task at hand. For example, for some vision tasks, we want to consider rotating or darkening an image, not just changing any old pixel.
- For fairness, we might want to show the ML model is fair on a given dataset and all unseen datasets that are "similar" (for some formal notion of "similar"). Training a recruiting tool to decide whom to interview on one population of applicants should ideally be fair on any future population. How can we specify these related distributions?

Verification task

- How do we check the available data for desired properties? For example, if we want to detect whether a dataset is fair or not, what should we be checking about the dataset?
- If we detect that the property does not hold, how do we fix the model, amend the property, or decide what new data to collect for retraining the model? What is the equivalent of a "counterexample" in the verification of an ML model and how do we use it?
- How do we exploit the explicit specification of unseen data to aid in the verification task?
- How can we extend standard verification techniques to operate over data distributions, perhaps taking advantage of the ways in which we formally specify unseen data?

These two key differences—inherent probabilistic nature of M and the role of data D—provide research opportunities for the formal methods community to advance specification and verification techniques for AI systems.

**2.3. Additional Formal Methods Opportunities**

Today's AI systems are developed with in mind to perform a particular task, e.g., face recognition or playing Go. How do we take into consideration the task that the deployed ML model is to perform in the specification and verification problem? For example, consider showing the robustness of a ML model, M, that does object recognition: For the task of identifying cars on the road, we would want M to be robust to the image of any car that has a dent in its side; but for the task of quality control in an automobile manufacturing line, we would not.

Section 2.2 focused on the verification task in formal methods. But the machinery of formal methods has also successfully been used recently for program synthesis [Gulwani, Polozov, and Singh 2017]. Rather than post-facto verification of a model M, can we develop a "correct-by-construction" approach in building M in the first place? For example, could we add the desired trustworthy property, P, as a constraint as we train and test M, with the intention of guaranteeing that P holds (perhaps for a given dataset or for a class of distributions) at deployment time?

Compositional reasoning enables us to do verification on large and complex systems. How does verifying a component of an AI system for a property "lift" to showing that property holds for the system? Conversely, how does one decompose an AI system into pieces, verify each with respect to a given property, and assert the property holds of the whole? Which properties are global (elude compositionality) and which are local? Decades of research in formal methods for compositional specification and verification give us a vocabulary and framework as a good starting point.

Statistics has a rich history in model checking[2] and model evaluation, using tools such as sensitivity analysis, prediction scoring, predictive checking, residual analysis, and model criticism. With the goal of validating an ML model satisfies a desired property, these statistical approaches can complement formal verification approaches, just as testing and simulation complement verification of computational systems. Even more relevantly, as mentioned in "The role of data" in Section 2.2, they can help with the evaluation of any statistical model used to specify unseen data, D, in the D, M ⊨ P problem. An opportunity for the formal methods community is to combine these statistical techniques with traditional verification techniques (for early work on such a combination, see [Younes and Simmons 2002]).

## 3. Promoting Trustworthy AI

Just as for trustworthy computing, formal methods is only one approach toward ensuring increased trust in AI systems. The community needs to explore many approaches to achieve trustworthy AI. Moreover, besides technical challenges, there are societal, policy, legal, and ethical challenges.

On October 30-November, 2020, Columbia University's Data Science Institute hosted an inaugural symposium on Trustworthy AI, sponsored by Capital One, a DSI industry affiliate. It brought together researchers from formal methods, security and privacy, fairness, and machine learning. Speakers from industry brought a reality check to the kinds of questions and approaches the academic community are pursuing. The participants identified research challenge areas, including:

- specification and verification techniques
- "correctness-by-construction" techniques
- new threat models and system-level adversarial attacks
- auditing processes that consider properties such as explainability, transparency, and responsibility
- ways to detect bias and de-bias data, machine learning algorithms, and their outputs
- systems infrastructure for experimenting for trustworthiness properties
- understanding the human element, e.g., where the machine is influencing human behavior
- understanding the societal element, including social welfare, social norms, morality, ethics, and law.

---

[2] Not to be confused with computer science's notion of model checking, where a finite state machine (computational model of a system) is checked against a given property specification [Clarke and Emerson 1980, Queille and Sifakis 1982].

In October 2019, the National Science Foundation announced a new program to fund [National AI Institutes](). One of the six themes is names "Trustworthy AI." It emphasizes properties such as reliability, explainability, privacy, and fairness.

Just as for trustworthy computing, government, academia and industry are coming together to drive a new research agenda in trustworthy AI. We are upping the ante on a holy grail!

**Acknowledgements**

In 2002-2003, I was fortunate to spend a sabbatical, hosted by Jim Larus, at Microsoft Research and witnessed firsthand how trustworthy computing permeated the company. I worked with Steve Lipner, Michael Howard, and Jon Pincus on defining and measuring the attack surface of different versions of Windows. It was also the year when the SLAM project [Ball et al. 2004], led by Tom Ball and Sriram Rajamani, showed how the use of formal methods could systematically detect bugs in device driver code, which at the time were responsible for a huge fraction of the infamous "blue screens of death." Whereas formal methods had already been shown to be useful and scalable for the hardware industry, the SLAM work was the first industry-scale project that showed the effectiveness of formal methods for software systems. I was also fortunate to be on the Microsoft Trustworthy Computing Academic Advisory Board, chaired by Fred Schneider and Deirdre Mulligan, from 2003-2007 and 2010-2012.

When I joined NSF in 2007 as the Assistant Director for the Computer and Information Science and Engineering Directorate, I promoted trustworthy computing across the directorate and with other federal agencies via NITRD. I would like to acknowledge my predecessor and successor CISE ADs, and all the NSF and NITRD program managers who cultivated the community in trustworthy computing. It is especially gratifying to see how the Trustworthy Computing program has grown to the Secure and Trustworthy Computing program, which continues to this day.

ACM sponsors the annual FAT* conference which originally promoted fairness, accountability, and transparency in machine learning. It has since grown to recognize other properties such as ethics, as well as the desirability of these properties for AI systems more generally.

I would like to acknowledge Shipra Agrawal, Roxana Geambasu, Daniel Hsu, and Suman Jana for their insights into what makes verifying AI systems different from verifying traditional computing systems. I thank Augustin Chaintreau for his recognition that at Columbia University, we have a team of like-minded individuals with different expertise who could come together and promote trustworthy AI. Other Columbia University team members include David Blei, Bo Cowgill, Bruce Kogut, Susan McGregor, Lalith Munansinghe (Barnard), Desmond Patton, Carl Vondrick, Eugene Wu, and Junfeng Yang. Final thanks to Roxana Geambasu and Tian Zheng who gave me comments on an earlier draft of this article.